\documentclass[lettersize,journal]{IEEEtran}
\usepackage{amsmath,amsfonts}
\usepackage{algorithmic}
\usepackage{algorithm}
\usepackage{array}
\usepackage[caption=false,font=normalsize,labelfont=sf,textfont=sf]{subfig}
\usepackage{textcomp}
\usepackage{stfloats}
\usepackage{url}
\usepackage{verbatim}
\usepackage{graphicx}
\usepackage{cite}
\usepackage{times}
\usepackage{epsfig}
\usepackage{amssymb}
\usepackage{arydshln}
\usepackage{bbding}
\usepackage{comment}
\usepackage{pifont}
\usepackage{booktabs}

\newcommand{\tabincell}[2]{\begin{tabular}{@{}#1@{}}#2\end{tabular}}

\usepackage{color}

\usepackage[pagebackref=true,breaklinks=true,letterpaper=true,colorlinks,bookmarks=false]{hyperref}

\newcommand{\cmark}{\ding{51}}
\newcommand{\xmark}{\ding{55}}

\begin{document}

\title{Masked Autoencoders in 3D Point Cloud Representation Learning}

\author{Jincen~Jiang,
        Xuequan~Lu,~\IEEEmembership{Senior Member,~IEEE,}
        Lizhi~Zhao,
        Richard~Dazaley,
        and~Meili~Wang,~\IEEEmembership{Member,~IEEE}
\thanks{J. Jiang, L. Zhao and M. Wang are with the College of Information Engineering, Northwest A\&F University, China (e-mail: jinec@nwafu.edu.cn; zhaolizhi@nwafu.edu.cn; wml@nwsuaf.edu.cn)}
\thanks{X. Lu is with the Department of Computer Science and IT, La Trobe University, Australia (e-mail: b.lu@latrobe.edu.au) }
\thanks{R. Dazaley is with the School of Information Technology, Deakin University, Australia (e-mail: richard.dazeley@deakin.edu.au) }
\thanks{Manuscript received August 8th, 2022; accepted September 10th, 2023.}}

\markboth{Journal of IEEE Transactions on Multimedia,~Vol.~xx, No.~yy, September~2023}%
{Shell \MakeLowercase{\textit{Jiang et al.}}: Masked Autoencoders in 3D Point Cloud Representation Learning}


\maketitle

\begin{abstract}
Transformer-based Self-supervised Representation Learning methods learn generic features from unlabeled datasets for providing useful network initialization parameters for downstream tasks.
Recently, methods based upon masking Autoencoders have been explored in the fields.          
The input can be intuitively masked due to regular content, like sequence words and 2D pixels. However, the extension to 3D point cloud is challenging due to irregularity. 
In this paper, we propose masked Autoencoders in 3D point cloud representation learning (abbreviated as MAE3D), a novel autoencoding paradigm for self-supervised learning. 
We first split the input point cloud into patches and mask a portion of them, then use our Patch Embedding Module to extract the features of unmasked patches. 
Secondly, we employ patch-wise MAE3D Transformers to learn both local features of point cloud patches and high-level contextual relationships between patches, then complete the latent representations of masked patches.
We use our Point Cloud Reconstruction Module with multi-task loss to complete the incomplete point cloud as a result. 
We conduct self-supervised pre-training on ShapeNet55 with the point cloud completion pre-text task and fine-tune the pre-trained model on ModelNet40 and ScanObjectNN (PB\_T50\_RS, the hardest variant). 
Comprehensive experiments demonstrate that the local features extracted by our MAE3D from point cloud patches are beneficial for downstream classification tasks, soundly outperforming state-of-the-art methods ($93.4\%$ and $86.2\%$ classification accuracy, respectively). 
\textit{Our source codes are available at: \href{https://github.com/Jinec98/MAE3D}{https://github.com/Jinec98/MAE3D}.}
\end{abstract}


\section{Introduction}
\label{sec:introduction}

Point cloud is a format of 3D data representation, which preserves the geometric information of 3D space, 
and is widely applied in autonomous driving, virtual reality, remote sensing and many other areas. In recent years, deep learning research on point clouds has developed rapidly, with promising results on tasks such as 3D point cloud shape classification and segmentation \cite{qi2017pointnet,dgcnn,yang2018foldingnet,qi2017pointnet++,2021pct,wu2019pointconv,wu20153d,2017o-cnn}. However, most existing methods require large labeled 3D point cloud datasets for supervised learning, which are expensive and time-consuming, driving the development of research on unsupervised point cloud learning. 

Self-supervised learning is a type of unsupervised learning, i.e., training a neural network with supervisory signals generated by the data itself \cite{SSLservey}.  
As a pioneer of Transformer-based self-supervised pre-training methods, BERT \cite{devlin2018bert} has made remarkable achievements in the field of natural language processing (NLP) by proposing the simple and effective masked language modeling pre-text task, which first randomly masks a portion of tokens within a text and then recovers the masked tokens by the  Transformer. 
Inspired by BERT, several self-supervised vision representation models have been designed. 
BEiT \cite{bao2021beit} introduces a masked image modeling task to pre-train the visual Transformer. They tokenize the original input image as discrete visual tokens, and input image patches (some patches are randomly masked) into the Transformer backbone to recover the tokens of masked patches. 
He \textit{et al.} \cite{he2021mae} presents the masked Autoencoders (MAE) method, which randomly masks the patches of the image and inputs the visible patches subset to the encoder to obtain the latent representations, which are then concatenated with the mask tokens and input to the decoder to reconstruct the missing pixels of the original input image. 

However, due to the gap between 3D point cloud data and image/text data,  BERT-style self-supervised pre-training models cannot be directly applied to point clouds.
Unlike text data with well-defined language vocabulary, point clouds do not have the concept of word in the NLP domain for the Transformer's input unit. 
In addition, unlike regular pixels, point clouds have irregular point distributions, and it is more challenging to extract contextual relationships between local patches than image patches with grid structure. 
Point-BERT \cite{yu2021pointbert} makes an attempt to apply the BERT-style model to self-supervised 3D representation learning by devising the Masked Point Modeling (MPM) task to pre-train the Transformer. They propose a point cloud Tokenizer to generate discrete point tokens (i.e., discrete integer number) and divide the point cloud into local patches, which are then randomly masked. The goal of the MPM task is to train the Transformer to recover the initial discrete point tokens of the masked local patches. 
However, since Point-BERT requires an additional pre-trained Tokenizer, their approach is relatively complicated and time-consuming. In addition, their model cannot learn high-level features directly without the help of MoCo \cite{moco} to enhance the feature extraction ability of Transformers.
More recently, some point cloud learning methods also take account of MAE. 
Point-MAE \cite{pang2022masked} uses Transformer based masked Autoencoder to reconstruct the point cloud.
They only use a fully connected layer as a prediction head to generate masked points and compute the reconstruction loss for every mask patch individually.
MaskPoint \cite{liu2022masked} trains a discriminator after the Transformer decoder to distinguish whether mask points are real or fake points, and underestimates the reconstruction of the entire point cloud.
Point-M2AE \cite{zhang2022point} conducts a reconstruction task with multi-stage masking, but they utilize a hierarchical U-Net instead of Transformer blocks to learn latent features at different levels.
Note that, all the above methods use a regular network, i.e., PointNet \cite{qi2017pointnet}, as patch embedding to extract the features of each patch. In addition, they use a simple linear layer to predict each local patch's coordinates discretely, ignoring the geometric relationship between each patch and the overall surface continuity of the point cloud during reconstruction.

\begin{figure*}[tbp]
\centering
\includegraphics[width=1\linewidth]{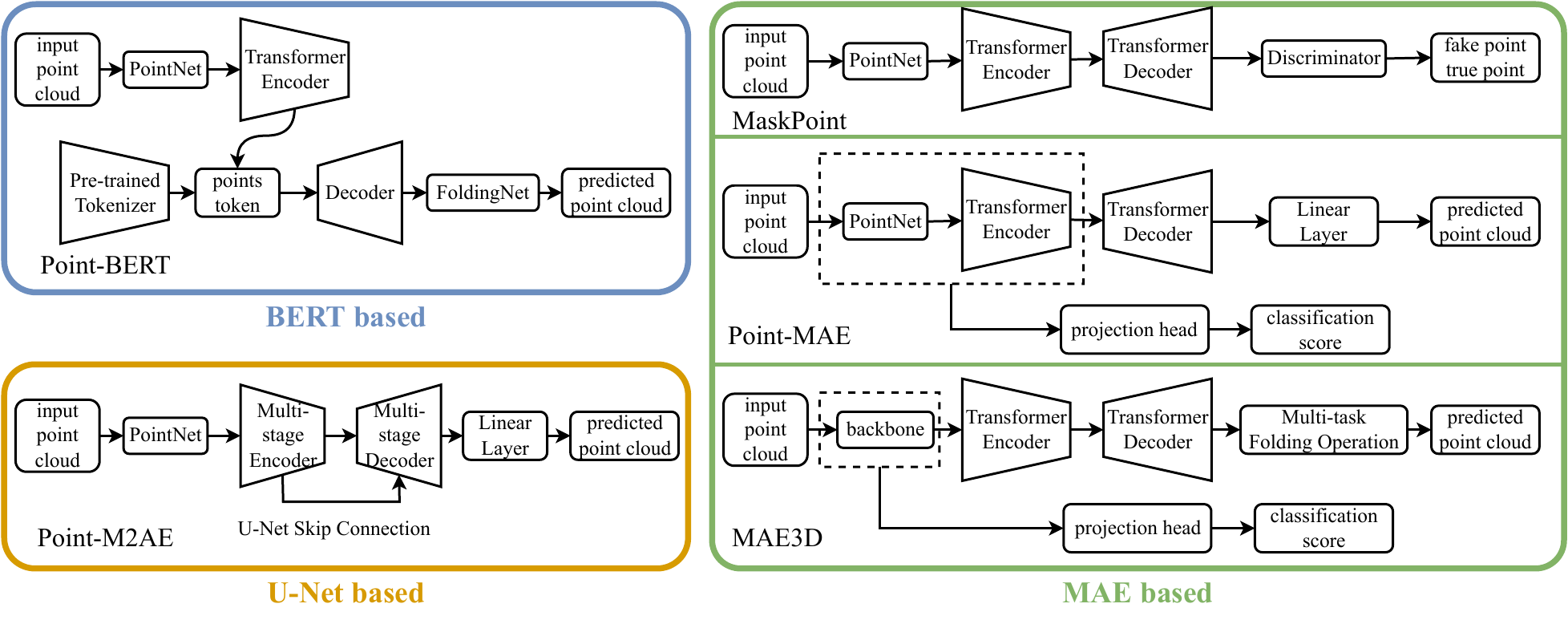}
\caption{Comparison of some recent works based on MPM pre-text task. We propose a masked Autoencoder based method (MAE3D) which is more general and lightweight and focuses more on point cloud reconstruction.
}
\label{fig:comparion}
\end{figure*}

Motivated by the above analysis, we propose 3D masked Autoencoders (MAE3D), a simple yet effective autoencoding paradigm for 3D point cloud self-supervised representation learning, to conduct the MPM pre-text task. 
We attempt to develop a general and lightweight framework making different point-based networks suit our paradigm. In this case, as shown in Figure \ref{fig:comparion}, we load only the backbone network in the downstream task without introducing additional parameters, i.e., the heavy transformer blocks. Further, we propose a multi-task folding operation, focusing more on the point cloud reconstruction task, which enables the network to learn both the patches relationship and the entire point cloud geometry structure. 
Specifically, our MAE3D first splits the input point cloud into patches by k-nearest neighborhood (KNN) algorithm, and mask a portion of these patches, which are subsequently completed. 
Our MAE3D consists of three main components: Patch Embedding Module, MAE3D Transformers and Point Cloud Reconstruction Module.
We devise the Patch Embedding Module with a patch feature extractor to extract latent representations of visible patches, which are concatenated with positional embedding and fed into MAE3D Transformers to complete the latent representations of masked patches. 
We then generate the entire point cloud global features by pooling the patch-wise latent representations and feed into our Point Cloud Reconstruction Module using our multi-task reconstruction loss to complete the incomplete point cloud.
After the pre-training phase, we load only the pre-trained patch feature extractor of Patch Embedding Module as our pre-trained model and evaluate it with 3D object classification as the downstream task.

The main contributions of this paper are:
\begin{enumerate}
    \item We propose MAE3D, a novel and effective Transformer-based self-supervised representation learning method with masked Autoencoder for 3D point cloud data, learning both local features of point cloud patches and the high-level contextual relationships between patches.

    \item Our MAE3D is lightweight and suitable for various point-based backbone networks. Only the pre-trained patch feature extractor is used for downstream tasks, without any additional parameters.

    \item We design a multi-task loss for reconstruction, enabling the network to learn both the geometry structure of local patches and the overall surface continuity of the entire point cloud.
    
    \item We conduct experiments and results show that our method pre-training on ShapeNet55 and fine-tuning on ModelNet40/ScanObjectNN soundly outperforms state-of-the-art techniques on the classification task, achieving $93.4\%$ and $86.2\%$ classification accuracy, respectively.
\end{enumerate}

\section{Related Work}
\label{sec:relatedwork}

\subsection{Supervised point cloud learning}
As a pioneer of point-based learning methods that directly consume raw point cloud representation without voxelization or projection, PointNet \cite{qi2017pointnet} leverages the symmetry of max-pooling to learn the permutation-invariance features of point clouds. 
PointNet++ \cite{qi2017pointnet++} devises a hierarchical architecture that recursively partitions the point cloud to extract local features more effectively, achieving better results than PointNet.
DGCNN \cite{dgcnn} is a graph-based method that creates a dynamic graph in the feature space and designs EdgeConv to learn the edge features of the graph in each layer.
Point Cloud Transformer (PCT) \cite{2021pct} applies the traditional Transformer \cite{attentionisallyouneed} to point cloud learning by constructing an order-invariant attention mechanism based on the Transformer.
Qiu \textit{et al.} \cite{qiu2021geometric} presented a network that takes into account both low-level geometric information of 3D space points directly and high-level local geometric context of feature space implicitly.

\subsection{Unsupervised point cloud Learning}
Yang \textit{et al.} propose FoldingNet \cite{yang2018foldingnet}, an end-to-end auto-encoder network for unsupervised learning of point clouds. They propose a novel folding-based decoder to transform the 2D grid onto a 3D surface represented by point cloud, achieving low reconstruction errors. 
MAP-VAE \cite{han2019mapvae} learns global and local  features of point clouds by combining global and local self-supervision.
LatentGAN \cite{latentgan} proposes a deep Autoencoder architecture to train a minimal GAN in the Autoencoder's latent space to learn point cloud representations.
Jiang \textit{et al.} \cite{jiang2021unsupervised} propose an unsupervised contrast learning method for point clouds.
They extract the feature representations of the original point cloud and its transformed version through a shared encoder.

\subsection{Pre-training Transformers}
Masked language model (MLM) proposed by BERT \cite{devlin2018bert} discards a portion of the input sequence and trains the model with supervised signals generated from the input sequence itself to recover the missing content. BERT has greatly contributed to the research of pre-training Transformers. 
Following BERT, BEiT \cite{bao2021beit} proposes the masked image modeling (MIM) task for self-supervised visual representation learning. 
BEiT first takes the discrete VAE \cite{ramesh2021zeroshot} as the image tokenizer, which tokenizes the image into discrete visual tokens. In the pre-training phase, BEiT randomly masks a portion of image patches, and feeds these corrupted patches into Transformer to learn to recover the visual tokens of the original image. 
More recently, MAE \cite{he2021mae} has proposed a more effective asymmetric encoder-decoder architecture for MIM tasks, where the encoder operates on the visible patches to extract the latent representation by a linear projection with concatenated positional embeddings. A lightweight decoder reconstructs the original image from the latent representation and mask tokens.

Yu \textit{et al.} \cite{yu2021pointbert} propose a BERT-style point cloud self-supervised representation learning model, named Point-BERT, to pre-train Transformer model through MPM task. 
The differences between Point-BERT and our MAE3D are significant: 
1) they require additional pre-training of a DGCNN-based Variation AutoEncoder (dVAE) and rely heavily on contrastive learning, while we directly use the patch embedding module to obtain the latent features of each patch and consider them as tokens, thus avoiding extra computational overhead. 
2) Their mask tokens are fed into the Transformer encoder, leading to early leakage of the location information, which is harmful to the latent feature of the network learning. By contrast, we replace the mask patches with a learnable mask token and take it as input to the Transformer decoder to force the encoder to focus on learning features from visible patches. 
3) They use a tokenizer to generate discrete integer point tokens as a self-supervised signal. We directly complete the latent representation of masked patches to reconstruct the point cloud, focusing more on the reconstruction task instead of predicting discrete point tokens.

We also elaborate the differences between some recent MAE-based works and our method: 1) Point-MAE \cite{pang2022masked} uses a simple linear layer as a prediction head to reconstruct the point cloud. By contrast, we use the folding operation, which can guarantee the continuity of the surface with better results. In addition, they only compute the loss for every mask patch, while our proposed multi-task loss first determines the location of each patch, then reconstructs their details by folding operation, and finally consolidates them to form a complete point cloud, ensuring the detail and continuity of the point cloud. 2) MaskPoint \cite{liu2022masked} trains the decoder to distinguish between the real and fake points instead of reconstructing the point cloud directly from latent features. 3) Point-M2AE \cite{zhang2022point} is not a Transformer-based network, but a hierarchical U-Net, which only learns features at different levels with masking strategy. Additionally, their multi-scale masking strategy reduces the mask region significantly after back-projecting, thus providing more input points, whereas we mask directly on the original input point cloud.

\section{Method}
\label{sec:method}
\begin{figure*}[tbp]
\centering
\includegraphics[width=1\linewidth]{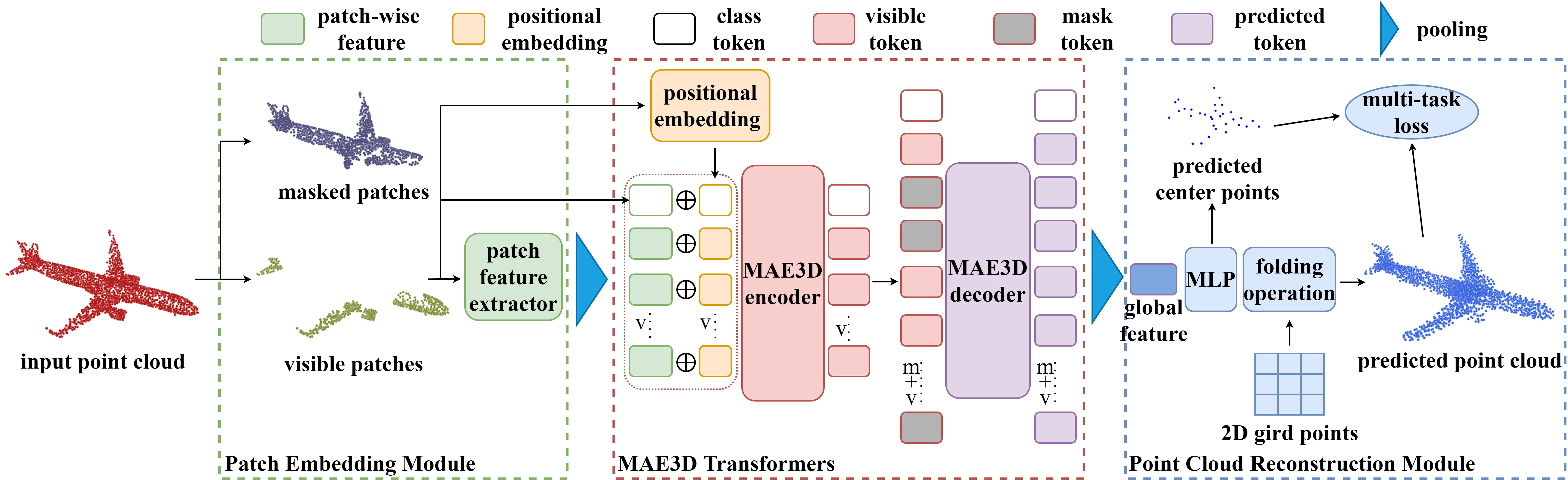}
\caption{Our MAE3D architecture. The input point cloud is split into patches and mask out a large portion of them. The remaining visible patches are fed to the patch feature extractor to obtain patch-wise features, adding positional embedding and class tokens, then passing them to MAE3D Transformers. The completed global features will be predicted by Transformers and used to reconstruct the point cloud. We develop a multi-task loss to measure both the predicted center points of patches and the reconstructed point cloud. 
}
\label{fig:overview}
\end{figure*}

Our MAE3D contains three main components: Patch Embedding Module, MAE3D Transformers, and Point Cloud Reconstruction Module. 
We first pre-train our model with MPM pre-text task, then fine-tune the pre-trained model with 3D object classification as the downstream task to verify its effectiveness. 
Figure \ref{fig:overview} shows the architecture of our work.

\subsection{Patch Embedding Module}
\label{sec::patch_embedding_moudle}
In this part, we demonstrate the Patch Embedding Module to map the geometric information of the point cloud into latent representations. 
3D point clouds are characterized by random order and irregular structure, and the irregularity of the input point clouds is aggravated by masking parts of the point cloud in the MPM pre-text task, which makes it difficult to obtain the embedding features by a simple linear projection as MAE \cite{he2021mae} and ViT \cite{vit} do for image patches.
To complete our MPM pretraining task, we first split point clouds into patches, then randomly mask a portion of them and feed these unmasked patches subset into our patch feature extractor to obtain the patch embedding features.

\textbf{Patch split.}
We treat the point cloud patch as the word of the BERT-style framework. Given a point cloud $P\in\mathbb{R}^{N\times3}$ with $N$ points, we first uniformly downsample it to a sampled point cloud $P_c = \{c_i \}_{i=1}^S\in\mathbb{R}^{S\times3}$ with $S$ points by farthest point sampling (FPS), where each point in $P_c$ is used as the center point of a patch.
For each point $c$ in $P_c$, we use the KNN algorithm to find a subset of local neighborhood centered on $c$ from $P$, and define all these subsets as point cloud patches $Q = \{P_i\}_{i=1}^S$. 
The FPS can generally distribute the center point of each patch evenly, which encourages these patches to cover all points of $P$ without duplication, i.e., $Q \approx P$. 

\textbf{Masking.}
We mask a part of the point cloud patches ($Q_M = \{P_i\}_{i \in m}$, $m$ is the sequence of the masked patch index) and take these remaining unmasked patches ($Q_V = \{P_i\}_{i \in v}$, $v$ is the sequence of visible patch index) as input. 
Specifically, we first randomly shuffle the list of patch indexes and remove that last part based on the masking ratio. The remaining patches are considered as visible patches which are then fed into the network.
There are two masking strategies: random masking and block masking. 
For random masking, we randomly and uniformly select a subset of $Q$ and remove it.
For block masking, we first select a random point $c_r$ in $P_c$, and calculate the K-nearest neighborhood of $c_r$ from $P_c$, then we remove patches in $Q$ corresponding to the point in that neighborhood.
Experimental results of these two strategies can be seen in Section \ref{sec:mae3d_completion_pretraining}. 

We completed our MPM pre-training task with a high masking ratio, i.e., $0.7$, making the task more difficult and challenging, to better learn the latent representations from the unmasked visible patches $Q_V$. Meanwhile, a more sparse point cloud input would also make our network more efficient.

An image can be spilt into a sufficient number of patches, and each of them contains enough pixel information \cite{he2021mae}. However, for point clouds, each patch has only a small number of points, e.g., $64$ points, and the number of split patches is very limited, e.g., only $32$ patches. 
Therefore, we choose the block masking strategy, which can preserve as much as possible the local spatial structure information of the remaining point cloud with a small number of points, which facilitates the network to obtain its latent representations.

\textbf{Patch feature extractor.}
Point-based networks, such as PointNet \cite{qi2017pointnet} and DGCNN \cite{dgcnn}, usually include a pooling layer to output global features of a point cloud. The former part before the pooling layer of these networks, i.e., a series of multi-layer perceptions (MLPs), can be naturally treated as the feature extractor $\Psi(\psi \mid Q)$ to obtain the latent representations of the input visible patch $\{\psi_i\}_{i \in v} = \Psi(Q_V)$ in our MAE3D framework. 
In this case, we can simply use the former part of any point-based network as our feature extractor. 
In this work, we take the individual patch-wise subsets as input, with the feature extractor capturing the local features of each patch.

\subsection{MAE3D Transformers}
\label{sec::mae3d_transformers}
MAE3D is in charge of recovering the latent representations of the missing area of the point cloud and predicting the whole point cloud information (i.e., point cloud global feature) with the limited input. Following the BERT-style manner, in this work, we use the standard Transformer that contains a series of Self-Attention layers and FNN blocks. We use an asymmetric structure, where the encoder focuses on only part of the point cloud (i.e., the visible patch), and the decoder reconstructs the whole point cloud global feature by encoded patches and mask tokens.

\textbf{Positional embedding.}
The positional embedding allows the operator in Self-Attention layers to well capture the contextual relationships of the input data. In image girds, the embedding of image patches is usually based on the sequence index using a trigonometric-based positional embedding method. However, in 3D data, the coordinates of each point naturally contain information about its position. Therefore, for each point cloud patch, we use its center point to represent the patch positional information with a trainable embedding layer. All patches' positional embedding can be defined as $ \Phi(c_i) = \{\phi_i\}_{i=1}^S $. 
The embedding function $\Phi$ is an MLP.

\textbf{MAE3D encoder.}
Our MAE3D encoder is formed by a series of Transformer blocks, and only applied to the visible patch $Q_V$. We concatenate the latent representations $\{\psi_i\}_{i \in v}$ obtained from each visible patch by the patch feature extractor and its positional embedding $\{\phi_i\}_{i \in v}$ together as the input embedding $\{x_i\}_{i \in v} = \{\textit{concat}(\psi_i, \phi_i)\}_{i \in v}$. Inspired by ViT \cite{vit}, we also put a class token $\Theta_0$ in front of the input sequence, which is formed by a random parameter of trainable vector $\psi_0$ and a randomly initialized positional embedding $\phi_0$, i.e., $\Theta_0 = \textit{concat}(\psi_0, \phi_0)$. 
The input to the MAE3D encoder can be denoted as $\chi = \{\Theta_0, x_i\}_{i \in v}$, which are then processed through a series of Transformer blocks. Similar to MAE \cite{he2021mae}, we discard the masked patches directly and do not use any mask tokens in the encoder part, which will save the computational time and memory effectively.

\textbf{MAE3D decoder.}
For the missing area, we first define mask tokens $\{\widetilde{\xi}_i\}_{i \in m}$ to represent the corresponding masked patch $Q_M$ that need to be predicted, which consist of a set of randomly initialized learnable latent representations $\{\widetilde{\psi}_i\}_{i \in m}$ and the positional embedding $\{\phi_i\}_{i \in m}$ of the missing patches. We provide the center point coordinates of each mask patch to compute the corresponding positional embedding, but the true locations of all points in each masked patch are not involved during the whole process. Similarly, we concatenate the initial latent representations and positional embedding to form the mask token as $\{\widetilde{\xi}_i\}_{i \in m} = \{\textit{concat}(\widetilde{\psi}_i, \phi_i)\}_{i \in m}$.

Our MAE3D encoder will output the encoded visible patches, called visible tokens $\{\xi_i\}_{i \in v}$. 
Since the MAE3D encoder and decoder have different dimensions, we use a linear layer to match their dimensions.
After that, we append a list of mask tokens at the end of visible tokens and unshuffle the entire token list (inverse of random shuffle) to align all tokens.
The input to the decoder consists of all patches tokens (i.e., both visible tokens and mask tokens) and class token, which can be defined as $\Xi = \{\Theta_0, \xi_i, \widetilde{\xi}_j\}_{i \in v, j \in m}$. The decoder has another series of Transformer blocks, and we simply use the standard Transformer following ViT \cite{vit} in image tasks. It is worth noting that during the whole process of our MAE3D Transformers, each point cloud patch will be viewed as an individual sample, i.e., MAE3D is a patch-wise approach. The output of the decoder will be a series of feature vectors representing each patch. We concatenate them in the point-wise dimension and then use a pooling layer to fuse the feature vectors into the global feature of the whole point cloud, which will be used for the pre-text task.

\subsection{Point Cloud Reconstruction Module}
\label{sec::point_cloud_reconstruction_module}
The Point Cloud Reconstruction Module is responsible for generating the output point cloud from the global feature. Folding-based methods, like FoldingNet \cite{yang2018foldingnet} and PCN \cite{yuan2018pcn},  utilize the folding operation to achieve this task. Motivated by these methods, we develop a patch-based strategy for our MAE3D approach, which can effectively reconstruct the point cloud.

\textbf{Patch center prediction.}
Similar to PCN \cite{yuan2018pcn}, we also adopt a series of MLPs to generate a coarse point cloud $\widetilde{P}_c$ as the first step. 
The MLPs enable better prediction of a sparse set of points to represent the approximate shape of the point cloud. 
A key point in our method is that this coarse point cloud contains a very small number of points, i.e., $32$ points, representing the center of each patch, which correspond to the $P_c$. 
Compared to previous works such as FoldingNet and PCN, which prioritize dense point cloud reconstruction but neglect structural information, our method introduces patch center prediction to improve the accuracy of the spatial location of each patch via the coarse point cloud output in the first stage, which leads to more faithful reconstructed results to the ground truth, depicting the more precise geometric shape of each patch and contextual relationships between patches.

\textbf{Folding-based patch deformation.}
In this part, for the predicted center of each patch, we use 2D gird points (i.e., $8$ rows $\times$ $8$ colons, totally have $64$ points) to generate each point in the patch with folding operation, and the number of each grid will exactly match the real patch. 
This folding operation takes the feature of the predicted center point and 2D grid as input and reconstructs a complete point cloud via deforming the grid.
With our improved strategy, we can reduce the computation and storage by avoiding the redundant points generated by PCN \cite{yuan2018pcn}, and the comparison between the predicted output $\widetilde{P}$ and ground truth $P$ is much more sensible.

\textbf{Reconstruction.}
Our MAE3D reconstructs a complete point cloud by predicting the features of each masked patch via inputting portion of the point cloud. With our Point Cloud Reconstruction Module, the predicted global feature of the point cloud can be reconstructed into a point cloud output. 
Our multi-task reconstruction loss function measures the Chamfer Distance (CD) between the reconstructed point cloud and the original one as well as the predicted center points of patches and the ground truth ones.
Unlike MAE \cite{he2021mae} that calculates loss only on masked patches, we will compute the loss on the entire point cloud. 
The Chamfer Distance can be defined as: 
\begin{align}
  \mathrm{CD}(S_1, S_2)=\frac{1}{|S_1|} \sum_{x \in S_1} \min _{y \in S_2}\|x-y\|_2^2 + 
            \frac{1}{|S_2|} \sum_{y \in S_2} \min _{x \in S_1}\|y-x\|_2^2,        
\end{align}
which calculates the average nearest point distance between the predicted point cloud $S_1$ and the ground truth point cloud $S_2$. 

Our multi-task loss function uses this metric twice to measure both predicted patch center $\widetilde{P}_c$ and predicted entire point cloud $\widetilde{P}$, which can be formulated as:
\begin{align}
  L = \mathrm{CD}(\widetilde{P_c}, P_c) + \alpha\mathrm{CD}(\widetilde{P}, P),
\end{align}
where $\alpha$ is the hyper-parameter representing the weight of the latter term.
This multi-task loss jointly uses the center point of each patch and the complete point cloud as self-supervised signal for point cloud completion.
Our method first locates each center point position of each patch, then reconstructs their details through the folding operation, and processes the entire point cloud. Thus, the center point prediction enables the network to learn the geometry relationship between each patch effectively. By contrast, Point-MAE \cite{pang2022masked} only predicts each masked patch without considering either the position of the center point of each patch or the continuity of the entire point cloud.

\subsection{Downstream Task}
\label{sec::downstram_task}
We use 3D object classification as our downstream task in this work to validate the performance of our MAE3D pre-text task pre-training. 
We take the former part of the backbone network (i.e., the patch feature extractor in Section \ref{sec::patch_embedding_moudle}) as the pre-trained model which contains valuable parameters.
In particular, the local feature and the spatial information of point cloud patches will be of great help for the downstream classification task.

We utilize two schemes to verify the capability of the pre-trained model. The first scheme is to use the pre-trained parameters to initialize the backbone and perform supervised training. The other one is to freeze the pre-trained model so that they will not be trained in the downstream task, and only train a linear classifier to classify the learned representations of this pre-trained model (i.e., self-supervised learning). We will demonstrate the classification results for these two validation schemes and the shape part segmentation task in Section \ref{sec:results}.

\section{Experimental Results}
\label{sec:results}

\subsection{Datasets}

\textbf{Completion pre-training.}
We use ShapeNet55 \cite{chang2015shapenet} as our pre-training dataset, which has $57,448$ models with $55$ categories, and all models will be used for the completion task to learn latent features. For each input point cloud, we divide them into $32$ point patches which have $64$ points for $2,048$ points inputs.
Then, we choose some patches to be masked with block or random mask sampling strategy. The rest of them, i.e., visible patches, will be used as input to the network.

\textbf{Object classification.}
We utilize ModelNet40 \cite{wu20153d} for 3D object classification. We follow the same data split protocols of previous methods, like PointNet \cite{qi2017pointnet} and DGCNN \cite{dgcnn}. The dataset contains $9,840$ models for training and $2,468$ models for testing, which consists of $40$ categories. We use $1,024$ points with $(x, y, z)$ normalized positions per model as the input, which is the same as previous works.

We also evaluate our method for the transfer learning on ScanObjectNN dataset \cite{uy-scanobjectnn-iccv19}, which contains $2,902$ unique object instances from $15$ categories. The objects are real-world point clouds based on the scanned indoor scene data. This dataset poses great challenges for the classification task, due to the involved background and incomplete data. We follow the official data split strategy, and conduct experiments on PB\_T50\_RS which is the most challenging dataset.

\textbf{Shape part segmentation.}
We verify our method on shape part segmentation using ShapeNet Part dataset \cite{yi2016scalable}, which contains $16,881$ shapes from $16$ categories. Each shape has $2$ to $6$ parts, consisting of $50$ specific part labels in total. We follow the official splits and point cloud sampling protocol in  \cite{chang2015shapenet}. Only the point coordinates are used as the input.

\subsection{Experimental Setting}
We use Adam optimizer for all our experiments and implement our work with PyTorch. Unlike DGCNN  \cite{dgcnn} that uses multiple GPUs, we only use a single TITAN RTX GPU for training. 
For the pre-training completion task, we use a batch size of $32$ for training. We use the dropouts of $0.5$ and the random seed $1$, which are the same as DGCNN. The learning rate is set to $0.0001$ under the cosine learning scheduler with the $0.0001$ weight decay. We train our model for $300$ epochs and choose the last checkpoint as our pre-trained model. 
For downstream 3D object classification on ModelNet40 \cite{wu20153d} and ScanObjectNN \cite{uy-scanobjectnn-iccv19}, we set the training batch size to $32$ for all experiments. The other settings follow DGCNN. Our method achieves comparable results without any voting strategy during testing. For fair comparison, we use the results without voting by other papers.

\subsection{MAE3D Completion Pre-training}
\label{sec:mae3d_completion_pretraining}
We experiment with the completion of 3D point clouds, which is the pre-text task for our method. Our MAE3D pre-training can be implemented simply and effectively. First, we remove a part of the point cloud by the block masking strategy and use the remaining point cloud as the input of the Patch Embedding Module. Then we predict the latent representation of missing parts by MAE3D Transformers, and finally, fuse the patch-wise features into the global feature to generate the point cloud output using the Point Cloud Reconstruction Module.

\begin{figure*}[htbp]
\centering
\begin{minipage}[b]{0.1\linewidth}
\begin{center}
Input
\end{center}
\end{minipage}
\begin{minipage}[b]{0.1\linewidth}
\begin{center}
Output
\end{center}
\end{minipage}
\begin{minipage}[b]{0.1\linewidth}
\begin{center}
Ground truth
\end{center}
\end{minipage}
\centering
\begin{minipage}[b]{0.1\linewidth}
\begin{center}
Input
\end{center}
\end{minipage}
\begin{minipage}[b]{0.1\linewidth}
\begin{center}
Output
\end{center}
\end{minipage}
\begin{minipage}[b]{0.1\linewidth}
\begin{center}
Ground truth
\end{center}
\end{minipage}\centering
\begin{minipage}[b]{0.1\linewidth}
\begin{center}
Input
\end{center}
\end{minipage}
\begin{minipage}[b]{0.1\linewidth}
\begin{center}
Output
\end{center}
\end{minipage}
\begin{minipage}[b]{0.1\linewidth}
\begin{center}
Ground truth
\end{center}
\end{minipage}\\
\centering
\begin{minipage}[b]{0.1\linewidth}
{\label{}\includegraphics[width=1\linewidth]{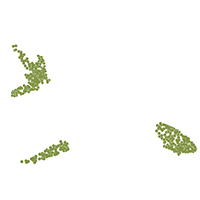}}
\end{minipage}
\begin{minipage}[b]{0.1\linewidth}
{\label{}\includegraphics[width=1\linewidth]{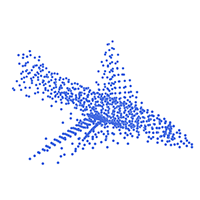}}
\end{minipage}
\begin{minipage}[b]{0.1\linewidth}
{\label{}\includegraphics[width=1\linewidth]{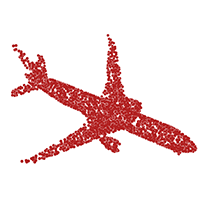}}
\end{minipage}
\begin{minipage}[b]{0.1\linewidth}
{\label{}\includegraphics[width=1\linewidth]{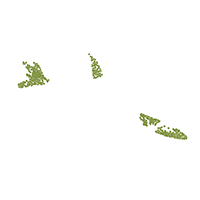}}
\end{minipage}
\begin{minipage}[b]{0.1\linewidth}
{\label{}\includegraphics[width=1\linewidth]{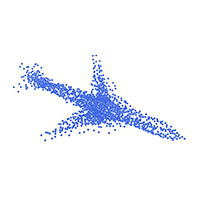}}
\end{minipage}
\begin{minipage}[b]{0.1\linewidth}
{\label{}\includegraphics[width=1\linewidth]{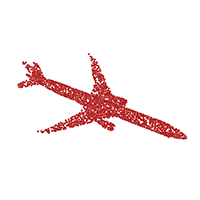}}
\end{minipage}
\begin{minipage}[b]{0.1\linewidth}
{\label{}\includegraphics[width=1\linewidth]{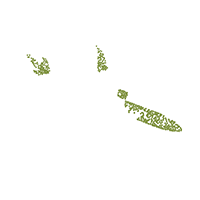}}
\end{minipage}
\begin{minipage}[b]{0.1\linewidth}
{\label{}\includegraphics[width=1\linewidth]{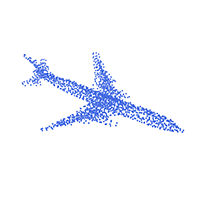}}
\end{minipage}
\begin{minipage}[b]{0.1\linewidth}
{\label{}\includegraphics[width=1\linewidth]{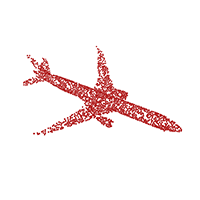}}
\end{minipage}\\
\centering
\begin{minipage}[b]{0.1\linewidth}
{\label{}\includegraphics[width=1\linewidth]{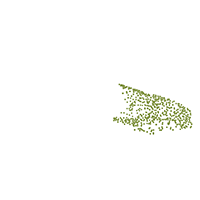}}
\end{minipage}
\begin{minipage}[b]{0.1\linewidth}
{\label{}\includegraphics[width=1\linewidth]{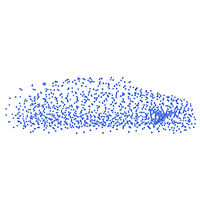}}
\end{minipage}
\begin{minipage}[b]{0.1\linewidth}
{\label{}\includegraphics[width=1\linewidth]{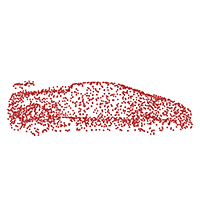}}
\end{minipage}
\begin{minipage}[b]{0.1\linewidth}
{\label{}\includegraphics[width=1\linewidth]{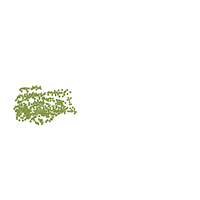}}
\end{minipage}
\begin{minipage}[b]{0.1\linewidth}
{\label{}\includegraphics[width=1\linewidth]{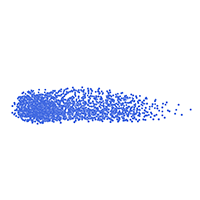}}
\end{minipage}
\begin{minipage}[b]{0.1\linewidth}
{\label{}\includegraphics[width=1\linewidth]{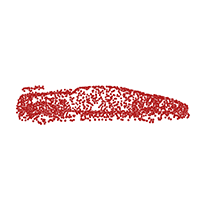}}
\end{minipage}
\begin{minipage}[b]{0.1\linewidth}
{\label{}\includegraphics[width=1\linewidth]{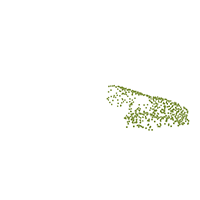}}
\end{minipage}
\begin{minipage}[b]{0.1\linewidth}
{\label{}\includegraphics[width=1\linewidth]{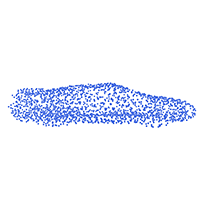}}
\end{minipage}
\begin{minipage}[b]{0.1\linewidth}
{\label{}\includegraphics[width=1\linewidth]{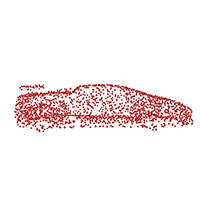}}
\end{minipage}\\
\centering
\begin{minipage}[b]{0.1\linewidth}
{\label{}\includegraphics[width=1\linewidth]{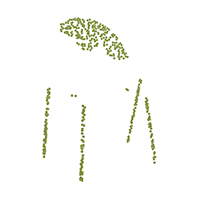}}
\end{minipage}
\begin{minipage}[b]{0.1\linewidth}
{\label{}\includegraphics[width=1\linewidth]{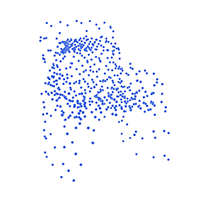}}
\end{minipage}
\begin{minipage}[b]{0.1\linewidth}
{\label{}\includegraphics[width=1\linewidth]{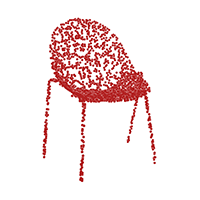}}
\end{minipage}
\begin{minipage}[b]{0.1\linewidth}
{\label{}\includegraphics[width=1\linewidth]{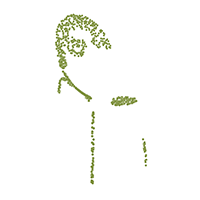}}
\end{minipage}
\begin{minipage}[b]{0.1\linewidth}
{\label{}\includegraphics[width=1\linewidth]{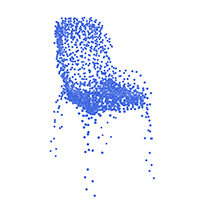}}
\end{minipage}
\begin{minipage}[b]{0.1\linewidth}
{\label{}\includegraphics[width=1\linewidth]{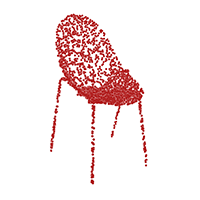}}
\end{minipage}
\begin{minipage}[b]{0.1\linewidth}
{\label{}\includegraphics[width=1\linewidth]{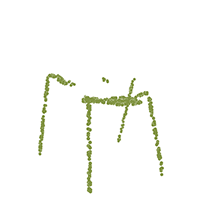}}
\end{minipage}
\begin{minipage}[b]{0.1\linewidth}
{\label{}\includegraphics[width=1\linewidth]{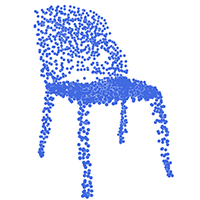}}
\end{minipage}
\begin{minipage}[b]{0.1\linewidth}
{\label{}\includegraphics[width=1\linewidth]{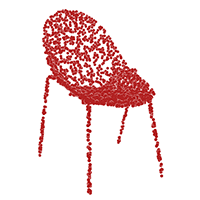}}
\end{minipage}\\
\centering
\begin{minipage}[b]{0.1\linewidth}
{\label{}\includegraphics[width=1\linewidth]{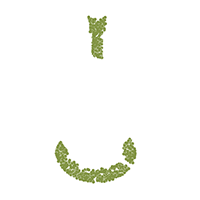}}
\end{minipage}
\begin{minipage}[b]{0.1\linewidth}
{\label{}\includegraphics[width=1\linewidth]{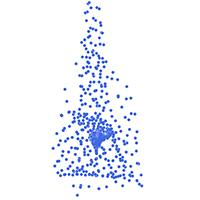}}
\end{minipage}
\begin{minipage}[b]{0.1\linewidth}
{\label{}\includegraphics[width=1\linewidth]{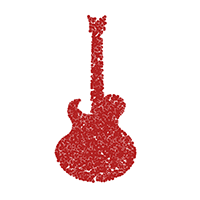}}
\end{minipage}
\begin{minipage}[b]{0.1\linewidth}
{\label{}\includegraphics[width=1\linewidth]{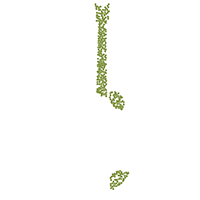}}
\end{minipage}
\begin{minipage}[b]{0.1\linewidth}
{\label{}\includegraphics[width=1\linewidth]{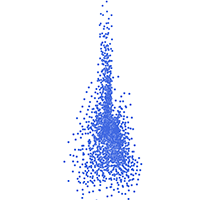}}
\end{minipage}
\begin{minipage}[b]{0.1\linewidth}
{\label{}\includegraphics[width=1\linewidth]{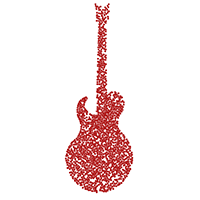}}
\end{minipage}
\begin{minipage}[b]{0.1\linewidth}
{\label{}\includegraphics[width=1\linewidth]{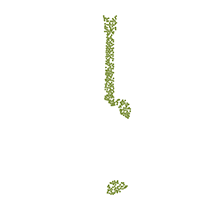}}
\end{minipage}
\begin{minipage}[b]{0.1\linewidth}
{\label{}\includegraphics[width=1\linewidth]{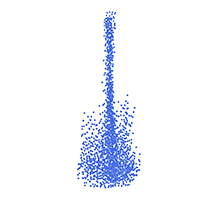}}
\end{minipage}
\begin{minipage}[b]{0.1\linewidth}
{\label{}\includegraphics[width=1\linewidth]{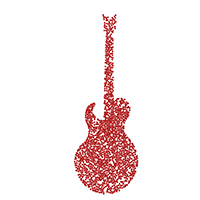}}
\end{minipage}\\
\centering
\begin{minipage}[b]{0.1\linewidth}
{\label{}\includegraphics[width=1\linewidth]{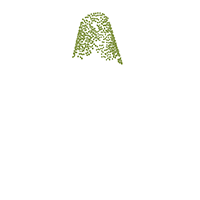}}
\end{minipage}
\begin{minipage}[b]{0.1\linewidth}
{\label{}\includegraphics[width=1\linewidth]{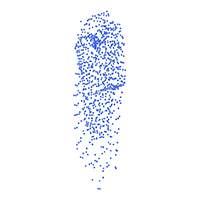}}
\end{minipage}
\begin{minipage}[b]{0.1\linewidth}
{\label{}\includegraphics[width=1\linewidth]{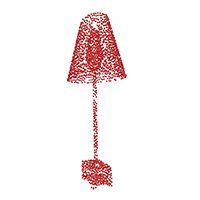}}
\end{minipage}
\begin{minipage}[b]{0.1\linewidth}
{\label{}\includegraphics[width=1\linewidth]{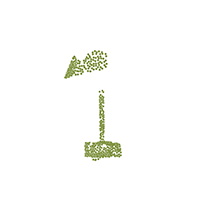}}
\end{minipage}
\begin{minipage}[b]{0.1\linewidth}
{\label{}\includegraphics[width=1\linewidth]{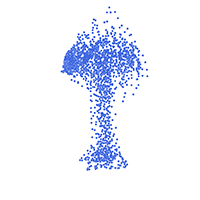}}
\end{minipage}
\begin{minipage}[b]{0.1\linewidth}
{\label{}\includegraphics[width=1\linewidth]{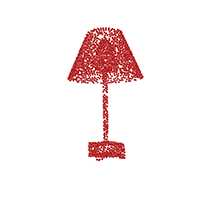}}
\end{minipage}
\begin{minipage}[b]{0.1\linewidth}
{\label{}\includegraphics[width=1\linewidth]{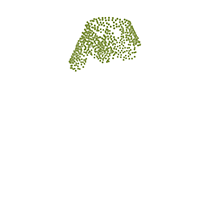}}
\end{minipage}
\begin{minipage}[b]{0.1\linewidth}
{\label{}\includegraphics[width=1\linewidth]{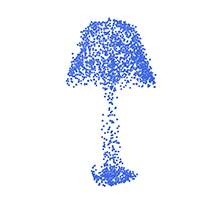}}
\end{minipage}
\begin{minipage}[b]{0.1\linewidth}
{\label{}\includegraphics[width=1\linewidth]{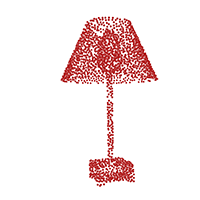}}
\end{minipage}\\
\centering
\begin{minipage}[b]{0.1\linewidth}
{\label{}\includegraphics[width=1\linewidth]{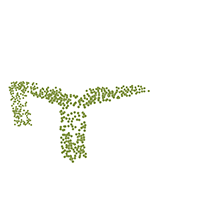}}
\end{minipage}
\begin{minipage}[b]{0.1\linewidth}
{\label{}\includegraphics[width=1\linewidth]{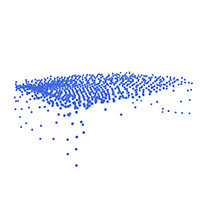}}
\end{minipage}
\begin{minipage}[b]{0.1\linewidth}
{\label{}\includegraphics[width=1\linewidth]{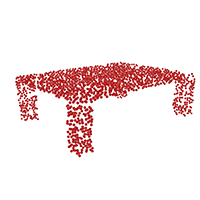}}
\end{minipage}
\begin{minipage}[b]{0.1\linewidth}
{\label{}\includegraphics[width=1\linewidth]{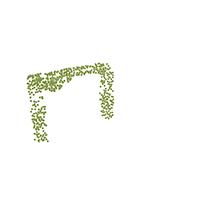}}
\end{minipage}
\begin{minipage}[b]{0.1\linewidth}
{\label{}\includegraphics[width=1\linewidth]{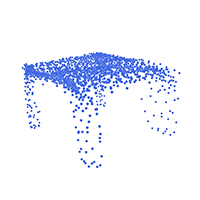}}
\end{minipage}
\begin{minipage}[b]{0.1\linewidth}
{\label{}\includegraphics[width=1\linewidth]{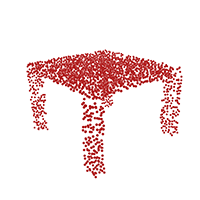}}
\end{minipage}
\begin{minipage}[b]{0.1\linewidth}
{\label{}\includegraphics[width=1\linewidth]{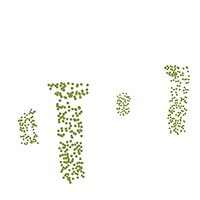}}
\end{minipage}
\begin{minipage}[b]{0.1\linewidth}
{\label{}\includegraphics[width=1\linewidth]{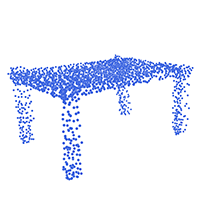}}
\end{minipage}
\begin{minipage}[b]{0.1\linewidth}
{\label{}\includegraphics[width=1\linewidth]{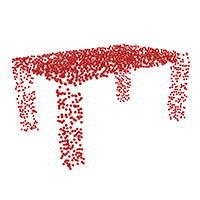}}
\end{minipage}\\
\centering
\begin{minipage}[b]{0.3\linewidth}
\begin{center}
FoldingNet \cite{yang2018foldingnet}
\end{center}
\end{minipage}
\begin{minipage}[b]{0.3\linewidth}
\begin{center}
PCN \cite{yuan2018pcn}
\end{center}
\end{minipage}
\begin{minipage}[b]{0.3\linewidth}
\begin{center}
\bf MAE3D
\end{center}
\end{minipage}\\
\caption{Comparison of sample completion results between our MAE3D and other SOTA methods with 0.7 block masking ratio.  }
\label{fig:completion_all}
\end{figure*}

We visualize the results of our method in Figure \ref{fig:completion_all}. It can be found that the output point cloud can be reconstructed well by our method with only inputting a small part of points. 
We also compare our method with some previous methods, like FoldingNet \cite{yang2018foldingnet} and PCN \cite{yuan2018pcn}, under the same settings. 
Since we do not have the test set in the pre-trained reconstruction task and should use the augmentation during the training stage, we provide the complete output and ground truth corresponding to the input of different transformations for each sample. 
Our MAE3D can accurately predict the masked patches, producing the best reconstructed point clouds among the compared methods. This  confirms that our MAE3D Transformers can obtain better local features of the point cloud patches and therefore our Point Cloud Reconstruction Module can further output more accurate point clouds.

We also calculate the Chamfer Distance of our MAE3D and some state-of-the-art (SOTA) methods,  illustrated in Table \ref{table:completion}. To provide a fair comparison, we re-implement and then train two most relevant methods (FoldingNet \cite{yang2018foldingnet} and PCN \cite{yuan2018pcn}) with the same masking strategy and masking ratio, i.e., the input are the same. It can be figured that our MAE3D can obtain the best results, which has the smallest error for the reconstructed output, and the pre-trained model also performs best in fine-tuning downstream task. In addition, we compared our MAE3D with Point-BERT \cite{yu2021pointbert}, and we use their reported results directly. It is worth mentioning that they input more points, i.e., a smaller masking ratio, and we still outperform them by $0.2\%$ on the downstream classification task.

\setlength{\tabcolsep}{4pt}
\begin{table}
\begin{center}
\caption{Comparison results of our method and SOTA techniques on the completion task for pre-training (Chamfer Distance) and the classification task for fine-tuning (classification accuracy). DGCNN is used as backbone in our Patch Embedding Module. }
\label{table:completion}
\begin{tabular}{l c c c c c}
\hline
\noalign{\smallskip}
Methods & \# Points & \tabincell{c}{Masking\\strategy} & \tabincell{c}{Masking\\ratio} & \tabincell{c}{CD\\($\times 10^{-3}$)} & Acc.\\
\noalign{\smallskip}
\hline
\noalign{\smallskip}
FoldingNet \cite{yang2018foldingnet}  & 1k & block & 0.7 & 6.858 & 90.4\\
PCN \cite{yuan2018pcn}  & 1k & block & 0.7 & 4.394 & 90.7\\
Point-BERT \cite{yu2021pointbert}  & 1k & block & [0.25, 0.45] & - & 93.2\\
Point-BERT \cite{yu2021pointbert}  & 1k & block & [0.55, 0.85] & - & 92.6\\
\bf{MAE3D} & 1k & block & 0.7 & \bf{3.127} & \bf{93.4}\\
\hline
\end{tabular}
\end{center}
\end{table}

\subsection{3D Object Classification}

\textbf{Pre-training evaluation.}
In this experiment, we select PointNet \cite{qi2017pointnet} and DGCNN \cite{dgcnn} as the backbone. After our MAE3D pre-training, we initialize the former part of the backbone (i.e., the feature extractor with a pooling layer) with the pre-trained model, which is used to obtain a global feature of the point cloud. The remaining classification branch (i.e., several MLPs) will be randomly initialized. We do not include the Transformer encoder in the downstream task network because it involves a series of blocks, which is a heavy structure, and including them affects the convergence of classification through experimental observation. The patch feature extractor merely is more lightweight and generalized to different point-based backbones.

As shown in Table \ref{table:supervised}, the pre-training evaluation based on MAE3D sees a significant improvement over the original backbone network, with an increase of $1.4\%$ on PointNet and $0.5\%$ on DGCNN.  It is worth noting that MAE3D achieves the best result of $93.4\%$ using DGCNN as the backbone, which is comparable with any other SOTA method. 
In addition, we perform experiments on  ModelNet10 and obtain the best results of $95.5\%$ (in Table \ref{table:supervised_mn10}), which exceeds all compared SOTA methods. 

\setlength{\tabcolsep}{4pt}
\begin{table}
\begin{center}
\caption{Classification comparison results of our method and supervised methods on ModelNet40. All results are without the voting strategy.}
\label{table:supervised}
\begin{tabular}{l c c c}
\hline
\noalign{\smallskip}
Methods & Pre-trained & \# Points & Acc.\\
\noalign{\smallskip}
\hline
\noalign{\smallskip}
PointNet++ \cite{qi2017pointnet++} & - & 1k & 90.7\\
HAPGN \cite{chen2020hapgn} & - & 1k & 91.7\\
PointCNN \cite{li2018pointcnn} & - & 1k & 92.5\\
PointConv \cite{wu2019pointconv} & - & 1k & 92.5\\
RSCNN \cite{liu2019relation} & - & 1k & 92.9\\
PCT \cite{2021pct} & - & 1k & 93.2\\
\hline
SO-Net \cite{li2018so} & \cmark & 2k & 90.9\\
3Dpatch \cite{sauder2019self} & \cmark & 1k & 92.4\\
SSC (RSCNN) \cite{chen2021shape} & \cmark & 1k & 93.0\\
Point-BERT \cite{yu2021pointbert} & \cmark & 1k & 93.2\\
Point-MAE \cite{pang2022masked} & \cmark & 1k & 93.2\\
Point-M2AE \cite{zhang2022point} & \cmark & 1k & \bf{93.4}\\
\hline
PointNet \cite{qi2017pointnet} & - & 1k & 89.2\\
MAE3D (PointNet) & \cmark & 1k & 90.6 (+1.4)\\
\hdashline
DGCNN \cite{dgcnn} & - & 1k & 92.9 \\
\bf{MAE3D (DGCNN)} & \cmark & 1k & \bf{93.4} (+0.5)\\
\hline
\end{tabular}
\end{center}
\end{table}

\setlength{\tabcolsep}{4pt}
\begin{table}
\begin{center}
\caption{Classification comparison results of our method and supervised methods on ModelNet10. }
\label{table:supervised_mn10}
\begin{tabular}{l c c}
\hline
\noalign{\smallskip}
Methods  & \# Points & Acc.\\
\noalign{\smallskip}
\hline
\noalign{\smallskip}
3D-GCN \cite{lin2020convolution} & 1k & 93.9\\
SO-Net \cite{li2018so} & 2k & 94.1\\
KCNet \cite{shen2018mining} & 1k & 94.4\\
PointCNN \cite{li2018pointcnn} & 1k & 94.9\\
Point2Sequence \cite{liu2019point2sequence} & 1k & 95.3\\
\bf{MAE3D} & 1k & \bf{95.5}\\
\hline
\end{tabular}
\end{center}
\end{table}

We also executed the classification experiment utilizing real-world data. This was conducted using all three variations found in ScanObjectNN: OBJ\_ONLY, OBJ\_BG, and PB\_T50\_RS. The results are presented in Table IV. From the results, it can be inferred that our pre-trained model garners an impressive result of 86.2\% with fine-tuning on one of the most challenging variants  PB\_T50\_RS. This performance surpasses both Point-MAE and PointMask by 1.0\% and 1.6\%, respectively. These results offer compelling evidence that our method demonstrates efficacy with real-world data.

\setlength{\tabcolsep}{4pt}
\begin{table}
\begin{center}
\caption{Classification comparison results of our method and supervised methods on ScanObjectNN.}
\label{table:scanobjectnn}
\begin{tabular}{l c c c}
\hline
\noalign{\smallskip}
Methods & OBJ\_ONLY & OBJ\_BG & PB\_T50\_RS\\
\noalign{\smallskip}
\hline
\noalign{\smallskip}
3DmFV \cite{ben20183dmfv}       & 73.8 & 68.2 & 63.0\\
PointNet \cite{qi2017pointnet}    & 79.2 & 73.3 & 68.2\\
SpiderCNN \cite{xu2018spidercnn}   & 79.5 & 77.1 & 73.7\\
PointNet++ \cite{qi2017pointnet++}  & 84.3 & 82.3 & 77.9\\
PointCNN \cite{li2018pointcnn}    & 85.5 & 86.1 & 78.5\\
Point-BERT \cite{yu2021pointbert}  & 88.1 & 87.4 & 83.1\\
PointMask \cite{liu2022masked} & 87.9 & 89.3 & 84.6\\
Point-MAE \cite{pang2022masked} & 88.3 & 90.0 & 85.2\\
Point-M2AE \cite{zhang2022point} & 88.8 & 91.2 & 86.4\\
\hdashline
DGCNN \cite{dgcnn}       & 86.2 & 82.8 & 78.1\\
\bf{MAE3D (DGCNN)} & 88.4 \textbf{(+2.2)} & 87.7  \textbf{(+4.9)}& 86.2 \textbf{(+8.1)}\\
\hline
\end{tabular}
\end{center}
\end{table}

\textbf{Few-shot learning.}
Following previous work \cite{sharma2020self}, we fine-tune our pre-trained model via few-shot learning. We also use the ``$K$-way $N$-shot" setting, where $K$ classes are randomly selected from the dataset, and then $(N+20)$ samples are chosen for each class. Therefore, there are $K\times(N+20)$ samples as the sub-dataset, with $K\times N$ samples as the training set and $K \times 20$ samples as the testing set. We adopt the dataset provided by PointBERT \cite{yu2021pointbert} to keep the same input as them. In our experiments, we also conduct $10$ independent experiments and calculate the mean and the standard deviation over these $10$ runs for $4$ schemes: ``$5$-way $10$-shot", ``$5$-way $20$-shot", ``$10$-way $10$-shot", and ``$10$-way $20$-shot". The results are shown in Table \ref{table:few_shot}, from which we can see that our MAE3D soundly outperforms PointBERT, indicating that our method is able to learn more discriminative features in a limited dataset.

\begin{table}
\begin{center}
\caption{Comparison results of few-shot learning on ModelNet40. The mean and standard deviation over 10 independent experiments are shown. The DGCNN's results are reported in PointBERT.}
\label{table:few_shot}
\resizebox{\columnwidth}{!}{
\begin{tabular}{l c c c c}
\hline
\noalign{\smallskip}
Methods & \tabincell{c}{5-way\\10-shot} & \tabincell{c}{5-way\\20-shot} & \tabincell{c}{10-way\\10-shot} & \tabincell{c}{10-way\\20-shot}\\
\noalign{\smallskip}
\hline
\noalign{\smallskip}
DGCNN (rand) \cite{yu2021pointbert}  & $91.8\pm3.7$ & $93.4\pm3.2$ & $86.3\pm6.2$ & $90.9\pm5.1$ \\
DGCNN (OcCo) \cite{yu2021pointbert}  & $91.9\pm3.3$ & $93.9\pm3.1$ & $86.4\pm5.4$ & $91.3\pm4.6$ \\
Point-BERT \cite{yu2021pointbert} &  $94.6\pm3.1$ & $96.3\pm2.7$ & $91.0\pm5.4$ & $92.7\pm5.1$ \\
Point-MAE \cite{pang2022masked}  &  $ \bf 96.3\pm2.5$ & $97.8\pm1.8$ & $ \bf 92.6\pm4.1$ & $95.0\pm3.0$ \\
\bf{MAE3D} & $95.2\pm3.1$ & $\bf 97.9\pm1.6$ & $ 91.1\pm4.6$  & $\bf 95.3\pm3.1$\\
\hline
\end{tabular}
}
\end{center}
\end{table}

\textbf{Linear classification evaluation.}
To verify that our pre-text task can effectively extract latent representations of point clouds, we use a linear classifier for 3D object classification, which follows the linear protocol on unsupervised learning. The pre-trained model is frozen and used to extract point cloud features, and only the linear classifier is trained. 

Table \ref{table:unsupervised} shows the comparison results of our method and the SOTA methods. For a fair comparison, we have classified the experimental results based on the pre-trained dataset. 
With the reconstruction as the pre-text task, our MAE3D significantly outperforms the other methods with pre-training on ShapeNet55, e.g., exceeding LatentGAN \cite{latentgan} by $6.8\%$ and exceeding FoldingNet  \cite{yang2018foldingnet} by $4.1\%$.

\setlength{\tabcolsep}{4pt}
\begin{table}
\begin{center}
\caption{Classification comparison results of our method and unsupervised methods on ModelNet40. ShapeNet55 and ModelNet40 pretrained cases are provided. }
\label{table:unsupervised}
\resizebox{\columnwidth}{!}{
\begin{tabular}{l c c c c}
\hline
\noalign{\smallskip}
Methods & \tabincell{c}{Pre-trained\\dataset} & \tabincell{c}{Pre-text task} & \# Points & Acc.\\
\noalign{\smallskip}
\hline
\noalign{\smallskip}
FoldingNet \cite{yang2018foldingnet} & ShapeNet55 & reconstruction & 2k & 88.4\\
VIPGAN \cite{han2019view} & ShapeNet55 & reconstruction & 2k & 90.2\\
LatentGAN \cite{latentgan} & ShapeNet55 & reconstruction & 2k & 85.7\\
PointCapsNet \cite{zhao20193d} & ShapeNet55 & reconstruction & 2k & 88.9\\
SSC (RSCNN) \cite{chen2021shape} & ShapeNet55 & shape self-correction & 2k & 92.4\\
\hdashline
\bf{MAE3D (DGCNN) }& ShapeNet55 & reconstruction & 2k & \bf{92.5}\\
\hline
FoldingNet \cite{yang2018foldingnet} & ModelNet40 & reconstruction & 2k & 84.4\\
LatentGAN \cite{latentgan} & ModelNet40 & reconstruction & 2k & 87.3\\
PointCapsNet \cite{zhao20193d} & ModelNet40 & reconstruction & 1k & 87.5\\
MAP-VAE \cite{han2019mapvae} & ModelNet40 & reconstruction & 2k & 90.2\\
Multi-task \cite{hassani2019unsupervised} & ModelNet40 & multiple & 2k & 89.1\\
PointHop \cite{zhang2020pointhop} & ModelNet40 & multiple & 1k & 89.1\\
GLR (RSCNN) \cite{rao2020global} & ModelNet40 & multiple & 1k & 92.2\\
\hdashline
\bf{MAE3D (DGCNN)} & ModelNet40 & reconstruction & 1k & \bf{92.4}\\
\hline
\end{tabular}
}
\end{center}
\end{table}

We also perform an experiment for training on limited data using linear classifier, and the results can be seen in Table \ref{table:limited_data}. Our MAE3D pre-trained model can achieve $88.3\%$ with DGCNN \cite{dgcnn} as backbone, even with only $20\%$ data for training. Furthermore, in some extreme cases, i.e., using only $1\%$ data, our method also achieves a good result of $61.7\%$, which exceeds FoldingNet \cite{yang2018foldingnet} by $5.3\%$, but is lower than $65.2\%$ of 3DPatch \cite{sauder2019self}. The reason is that MAE3D removes lots of point cloud patches (with $0.7$ masking) during the pre-training, and only learns the features of a limited number of points. When processing the downstream classification task, a very small amount of data will greatly affect the data diversity and result in less performance improvement.

\setlength{\tabcolsep}{4pt}
\begin{table}
\begin{center}
\caption{Comparison of 3D object classification accuracy results with limited training data (different ratios). The results without being reported previously are marked as `-'. }
\label{table:limited_data}
\begin{tabular}{l c c c c c}
\hline
\noalign{\smallskip}
Methods & 1\% & 2\% & 5\% & 10\% & 20\%\\
\noalign{\smallskip}
\hline
\noalign{\smallskip}
FoldingNet \cite{yang2018foldingnet}  & 56.4 & 66.9 & 75.6 & 81.2 & 83.6\\
3DPatch \cite{sauder2019self}  & \bf{65.2} & - & - & 84.4 & -\\
\bf{MAE3D} & 61.7 & \bf{69.2} & \bf{80.8} & \bf{84.7} & \bf{88.3}\\
\hline
\end{tabular}
\end{center}
\end{table}

\subsection{Shape Part Segmentation}
\label{sec:partseg}
We also conduct another downstream task, i.e., shape part segmentation, to further verify our method's effectiveness.
We use the mean Intersection-over-Union (mIoU) of instances as the evaluation metric. The results are reported in Table \ref{table:partseg} which shows our MAE3D using DGCNN as the backbone improves over the original DGCNN by $0.4\%$ in terms of mIoU of instances.
Though our downstream network has no additional parameters compared to the backbone, i.e., no heavy-weight Transformer blocks, we achieve comparable results to the recent Transformer-based methods like Point-BERT \cite{yu2021pointbert}.

\setlength{\tabcolsep}{3pt}
\begin{table*}[htb]\footnotesize
    \caption{Shape part segmentation results of our method and state-of-the-art techniques on ShapeNet Part dataset.}
    \label{table:partseg}
    \centering
    \begin{tabular}{l c c c c c c c c c c c c c c c c c c}
        \hline
         Methods & \tabincell{c}{Instance\\mIOU} & air. & bag & cap & car & chair & ear. & guit. & kni. & lam. & lap. & mot. & mug & pist. & rock. & ska. & tab.\\
         \hline
Kd-Net \cite{klokov2017escape} & 82.3 & 80.1 & 74.6 & 74.3 & 70.3 & 88.6 & 73.5 & 90.2 & 87.2 & 81.0 & 84.9 & 87.4 & 86.7 & 78.1 & 51.8 & 69.9 & 80.3\\
PointNet \cite{qi2017pointnet} & 83.7 & 83.4 & 78.7 & 82.5 & 74.9 & 89.6 & 73.0 & 91.5 & 85.9 & 80.8 & 95.3 & 65.2 & 93.0 & 81.2 & 57.9 & 72.8 & 80.6\\
SO-Net \cite{li2018so} & 84.9 & 82.8 & 77.8 & 88.0 & 77.3 & 90.6 & 73.5 & 90.7 & 83.9 & 82.8 & 94.8 & 69.1 & 94.2 & 80.9 & 53.1 & 72.9 & 83.0\\
PointNet++ \cite{qi2017pointnet++} & 85.1 & 82.4 & 79.0 & 87.7 & 77.3 & 90.8 & 71.8 & 91.0 & 85.9 & 83.7 & 95.3 & 71.6 & 94.1 & 81.3 & 58.7 & 76.4 & 82.6\\
3D-GCN \cite{lin2020convolution} & 85.1 & 83.1 & 84.0 & 86.6 & 77.5 & 90.3 & 74.1 & 90.9 & 86.4 & 83.8 & 95.6 & 66.8 & 94.8 & 81.3 & 59.6 & 75.7 & 82.8\\
Point-BERT \cite{yu2021pointbert} & 85.6 & 84.3 & 84.8 & 88.0 & 79.8 & 91.0 & 81.7 & 91.6 & 87.9 & 85.2 & 95.6 & 75.6 & 94.7 & 84.3 & 63.4 & 76.3 & 81.5\\
MaskPoint \cite{liu2022masked} & 86.0 & 84.2 & 85.6 & 88.1 & 80.3 & 91.2 & 79.5 & 91.9 & 87.8 & 86.2 & 95.3 & 76.9 & 95.0 & 85.3 & 64.4 & 76.9 & 81.8\\
Point-MAE \cite{pang2022masked} & 86.1 & 84.3 & 85.0 & 88.3 & 80.5 & 91.3 & 78.5 & 92.1 & 87.4 & 86.1 & 96.1 & 75.2 & 94.6 & 84.7 & 63.5 & 77.1 & 82.4\\
\hdashline
DGCNN \cite{dgcnn} & 85.2 & 84.0 & 83.4 & 86.7 & 77.8 & 90.6 & 74.7 & 91.2 & 87.5 & 82.8 & 95.7 & 66.3 & 94.9 & 81.1 & 63.5 & 74.5 & 82.6\\
MAE3D (DGCNN) & 85.6 \bf{(+0.4)} & 83.7 & 81.7 & 85.3 & 78.5 & 90.8 & 80.1 & 91.4 & 89.1 & 84.6 & 95.5 & 66.7 & 94.8 & 81.9 & 58.5 & 74.6 & 82.7\\
    \hline
    \end{tabular}
\end{table*}

\subsection{Ablation Studies}

\begin{figure}[htbp]
\centering
\includegraphics[width=1.0\linewidth]{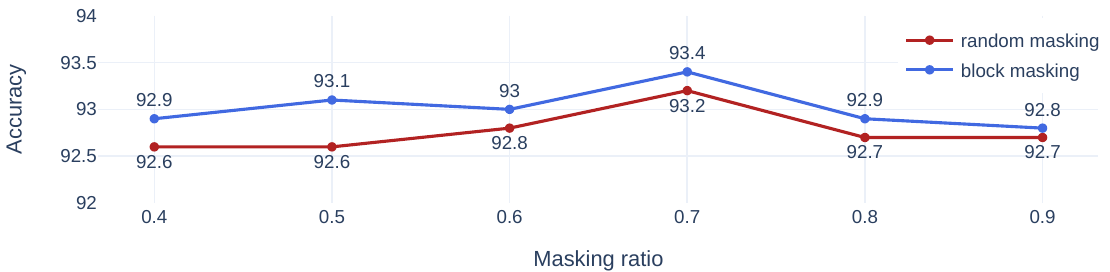}
\caption{Comparison of different masking strategies and masking ratios on downstream classification task (with DGCNN as the backbone). The y-axis is ModelNet40 testing accuracy (\%). }
\label{fig:masking}
\end{figure}

\textbf{Masking.}
In this part, we show the results of two masking strategies with different masking ratios in Figure \ref{fig:masking}. It can be observed that using block masking can obtain better results in all cases, which has higher accuracy on the classification task. Besides, a high masking ratio, i.e., $0.7$, works well on both block masking and random masking. It suggests that the challenge caused by a high masking ratio will benefit the network to capture the latent representations of the sample, which facilitates the classification results. We also calculated the Chamfer Distance for the reconstruction of these two strategies on the ShapeNet55 dataset \cite{chang2015shapenet} with the masking ratio of $0.7$, and the results are shown in Table \ref{table:masking_strategies} (cases A and B). It can be seen that block masking still performs better than random masking.

\textbf{Loss.}
To verify the effectiveness of our proposed multi-task loss, we conduct an experiment on whether using center point loss. As shown in Table \ref{table:masking_strategies} (cases C and D), the prediction of the center point for each patch serves a critical role, allowing the network to capture the geometric position relationship between each patch, which contributes to better results in the reconstruction of the entire point clouds, i.e., a smaller Chamfer Distance error.

\setlength{\tabcolsep}{4pt}
\begin{table}
\begin{center}
\caption{Comparison results of random/block masking and the different loss options. The Chamfer Distance is calculated on ShapeNet55 for pre-training with 0.7 masking ratio. The classification Accuracy is calculated on ModelNet40. 
}
\label{table:masking_strategies}
\begin{tabular}{l c c c c c}
\hline
\noalign{\smallskip}
Cases & Masking Strategies & Loss & CD ($\times 10^{-3}$) & Acc.\\
\noalign{\smallskip}
\hline
\noalign{\smallskip}
A & random & multi-task loss & 3.260 & 93.2\\
B & block & multi-task loss & \bf{3.127} & \bf{93.4}\\
\hdashline
C & block & w/o center point & 4.034 & 93.0\\
D & block & multi-task loss & \bf{3.127} & \bf{93.4}\\
\hline
\end{tabular}
\end{center}
\end{table}

\textbf{Patch size.}
We perform the ablation study on the number of points in each patch (i.e., patch size). A large patch size causes a large overlap of each patch in the split, which prevents the selected points from covering as many points as possible in the original point cloud. Fewer points per patch will affect the effectiveness of the patch feature extractor. In general, it is more difficult to learn features with fewer points as input. As shown in Table \ref{table:patch_size}, $64$ points per patch is the best.

\setlength{\tabcolsep}{4pt}
\begin{table}
\centering
\caption{Comparison results of different patch sizes. ShapeNet55 is used for pre-training with block masking and ratio 0.7. ModelNet40 is used for fine-tuning with DGCNN as backbone. 
}
  \begin{tabular}{l|cccc}
     \hline
     \#. patches & 16 & 32 & 64 & 128\\
     \#. points per patch & 128 & 64 & 32 & 16\\
     \hline
     Acc.(\%) & 92.9 & \textbf{93.4} & 93.2 & 93.1\\
     \hline
  \end{tabular}
\label{table:patch_size}
\end{table}

\textbf{Decoder design.}
We also conduct a study on our MAE3D decoder. Table \ref{table:decoder} studies the decoder depth (number of Transformer blocks) and decoder width (number of channels). The results indicate that 6 blocks and 1024 channels achieve the best performance.

\setlength{\tabcolsep}{4pt}
\begin{table}
\vspace{0.3cm}
\centering
\caption{Comparison results of the decoder structure. The decoder is used for the completion task. ShapeNet55 is used for pre-training, and ModelNet40 is used for fine-tuning. The classification accuracy on ModelNet40 is used here. 
}
  \begin{tabular}{l|cccc}
     \hline
     \#. blocks & 2 & 4 & 6 & 8\\
     \hline
     Acc.(\%) & 93.3 & 92.9 & \textbf{93.4} & 92.9\\
     \hline
  \end{tabular}
  \vspace{0.5cm}
  \begin{tabular}{l|ccc}
     \hline
     \#. channels & 512 & 1024 & 2048\\
     \hline
     Acc.(\%) & 92.7 & \textbf{93.4} & 92.8\\
     \hline
 \end{tabular}
\label{table:decoder}
\end{table}

\begin{figure}[htbp]
\centering
\includegraphics[width=1.0\linewidth]{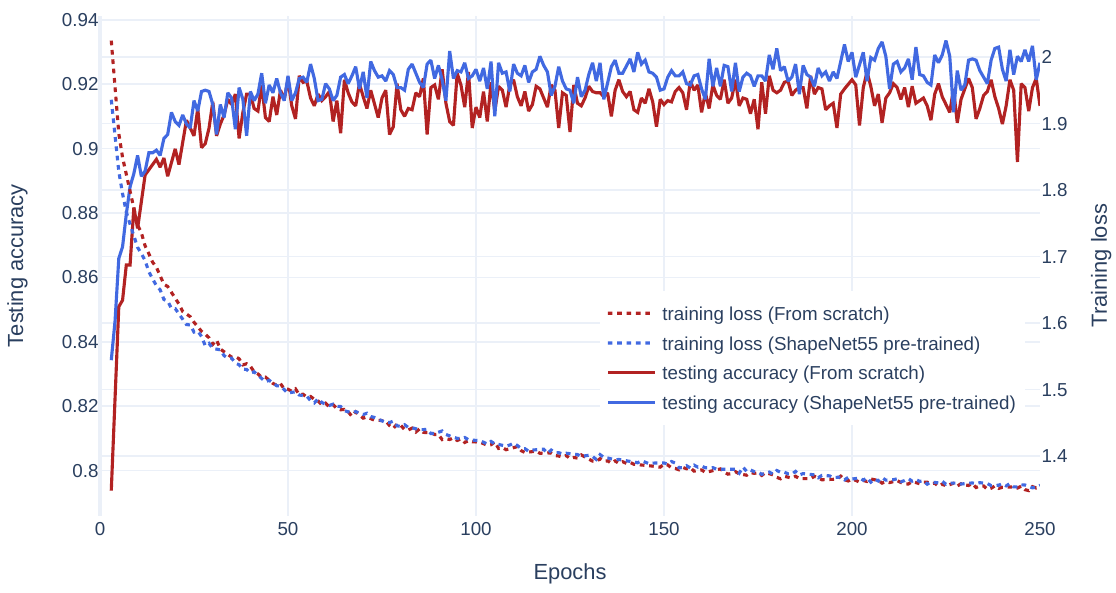}
\caption{Comparison with or without using pre-trained model for ModelNet40. The y-axes on the left/right are testing accuracy (\%) and training loss, respectively. }
\label{fig:pretraining}
\end{figure}

\textbf{Training with v.s. without the pre-trained model.}
Our pre-trained model can effectively improve the capability of the backbone. It is interesting to find that using our pre-trained parameters as initialization and then fine-tuning on downstream tasks can enable the model to capture the point cloud features at the beginning of the training process. As shown in Figure \ref{fig:pretraining}, the case with pre-training achieves better classification performance than training from scratch (i.e., original DGCNN \cite{dgcnn}). The model can converge to the optimal solution more efficiently and accurately.

\textbf{Pre-training with v.s. without MAE3D Transformers.}
We examine whether our MAE3D Transformers can predict the latent representations of the masked patches, i.e., a more complete and significant global feature of the point cloud can be obtained by MAE3D Transformers. We perform a comparison experiment to demonstrate this. As shown in Table \ref{table:transformers}, the network pre-training with MAE3D Transformers can achieve a better result, with a $1\%$ improvement compared to the network without it.

\setlength{\tabcolsep}{4pt}
\begin{table}
\begin{center}
\caption{Comparison results of whether using MAE3D Transformers during pre-training. ShapeNet55 is used for pre-training with block masking and ratio 0.7. 
}
\label{table:transformers}
\begin{tabular}{l c}
\hline
\noalign{\smallskip}
Cases & Acc.\\
\noalign{\smallskip}
\hline
\noalign{\smallskip}
Without MAE3D Transformers & 92.4\\
With MAE3D Transformers & \bf{93.4}\\
\hline
\end{tabular}
\end{center}
\end{table}

\textbf{Fine-tuning with v.s. without Transformer Encoder.}
A significant contribution of our method is its ability to fine-tune downstream tasks using only the backbone network, without the complex transformer encoder, making our network more lightweight. 
As shown in Table \ref{table:transformers_encoder}, Point-MAE \cite{pang2022masked} requires loading the transformer encoder, which contains $22.0 M$ parameters ($12$ times more than ours of $1.8 M$), but we achieve improved results with a much more lightweight model.

For more comprehensive comparison, we reproduce Point-MAE by removing the transformer encoder but obtain subpar results. 
In contrast, our method can yield comparable results when loading the heavy transformer encoder. 
We hypothesize that the transformer encoder is focused on predicting the masked patches feature to reconstruct the point cloud in the pre-text task, while the backbone network primarily extracts the point cloud features. Without loading the transformer encoder, the learned point cloud features will not be excessively transformed by a complex network, resulting in a more intuitive point cloud feature representation.

\setlength{\tabcolsep}{4pt}
\begin{table}
\begin{center}
\caption{Comparison results of whether using Transformer Encoder during fine-tuning.}
\label{table:transformers_encoder}
\begin{tabular}{l c r c}
\hline
\noalign{\smallskip}
Methods & Transformer Encoder & Params. & Acc.\\
\noalign{\smallskip}
\hline
\noalign{\smallskip}
Point-MAE \cite{pang2022masked} & \cmark & 22.00 M & 93.2 \\
Point-MAE \cite{pang2022masked} & \xmark & 0.67 M & 92.5\\
MAE3D & \cmark & 40.65 M & 93.2 \\
MAE3D & \xmark & 1.80 M & \textbf{93.4}\\
\hline
\end{tabular}
\end{center}
\end{table}

\textbf{Comparison with different feature extractors.}
We perform an experiment comparing the backbone network used as the feature extractor in our model. For a fair comparison with Point-MAE \cite{pang2022masked}, we employ PointNet \cite{qi2017pointnet} as the backbone network to extract features from the input points. Simultaneously, we reproduce Point-MAE by altering their encoder with DGCNN \cite{dgcnn} and carrying out the complete pre-training and fine-tuning process. The results are displayed in Table \ref{table:feature_extractor}, which indicates that our method significantly outperforms Point-MAE by 0.3\% when using DGCNN. We also achieve close results when PointNet is used as the backbone. It's noteworthy that Point-MAE incorporates Transformer blocks in the encoder during fine-tuning, which aids the simple PointNet in extracting more comprehensive information.

\setlength{\tabcolsep}{4pt}
\begin{table}
\begin{center}
\caption{Comparison with different feature extractors.}
\label{table:feature_extractor}
\begin{tabular}{l c c}
\hline
\noalign{\smallskip}
Methods & Feature extractor & Acc.\\
\noalign{\smallskip}
\hline
\noalign{\smallskip}
PointNet \cite{qi2017pointnet} & - & 89.2 \\
Point-MAE \cite{pang2022masked} & PointNet + Transformer & 93.2 \\
MAE3D & PointNet & 90.6\\
\hdashline
DGCNN \cite{dgcnn} & - & 92.9\\
Point-MAE \cite{pang2022masked} & DGCNN + Transformer & 93.1 \\
MAE3D & DGCNN & \textbf{93.4}\\
\hline
\end{tabular}
\end{center}
\end{table}

\subsection{Visualization}
\label{sec:visualization}

\begin{figure}[htbp]
\centering
\includegraphics[width=1\linewidth]{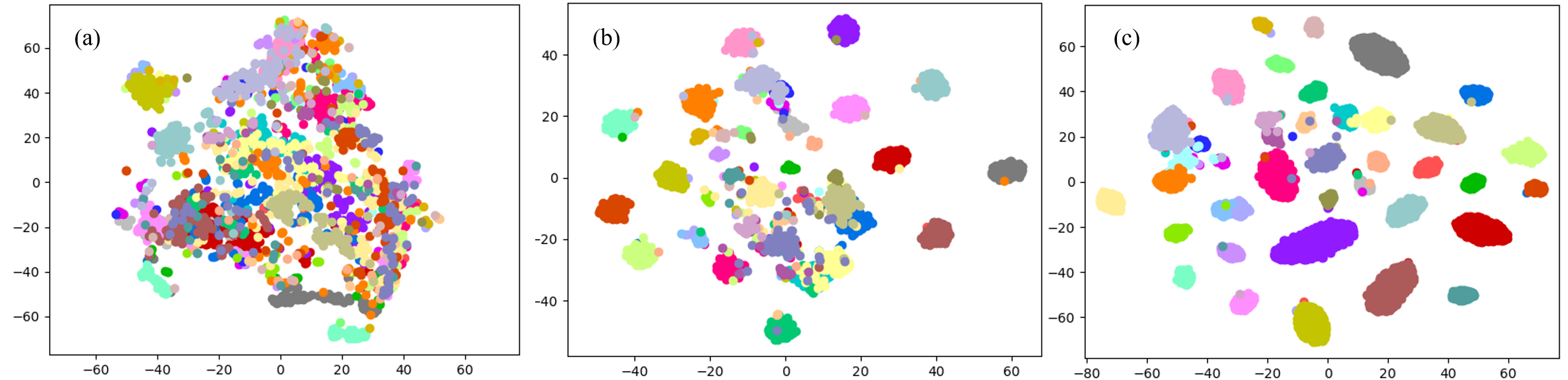}
\caption{T-SNE visualization of features learned by MAE3D and original DGCNN. (a) MAE3D pre-training, (b) original DGCNN, (c) MAE3D fine-tuning. }
\label{fig:tsne}
\end{figure}

In Figure \ref{fig:tsne}, we visualize the learned features on ModelNet40 \cite{wu20153d} with t-SNE visualization. It can be seen that our MAE3D can separate  different features relatively well even in the pre-training process, which indicates that our MAE3D Transformers can learn local features among point cloud patches without specific labels. It is interesting to observe that comparing with the original DGCNN \cite{dgcnn} (i.e., training from scratch), our MAE3D could enable more separable features after fine-tuning, which implies that our pre-trained model can effectively facilitate the downstream task.

\section{Conclusion}
\label{sec:conclusion}
In this paper, we proposed MAE3D for point cloud self-supervised representation learning through the MPM pre-text task, which can learn the discriminative features of point cloud patches and the high-level contextual relationships between patches. 
After pre-training, we load the pre-trained patch feature extractor and fine-tune it on the downstream task of 3D object classification, achieving state-of-the-art accuracy of $93.4\%$ on ModelNet40 and $86.2\%$ on ScanObjectNN (PB\_T50\_RS). This demonstrates the superiority of our proposed method over state-of-the-art methods.

\section*{Acknowledgments}
This work was supported by the National Key Research and Development Program of China (Grant Number 2022YFD1300200) and the Shaanxi Province Key Research and Development Program (Grant Number 2022QFY11-03).

\bibliographystyle{IEEEtran}
\bibliography{IEEEabrv,egbib}

\newpage

\section{Biography Section}

\begin{IEEEbiography}[{\includegraphics[width=1in,height=1.25in,clip,keepaspectratio]{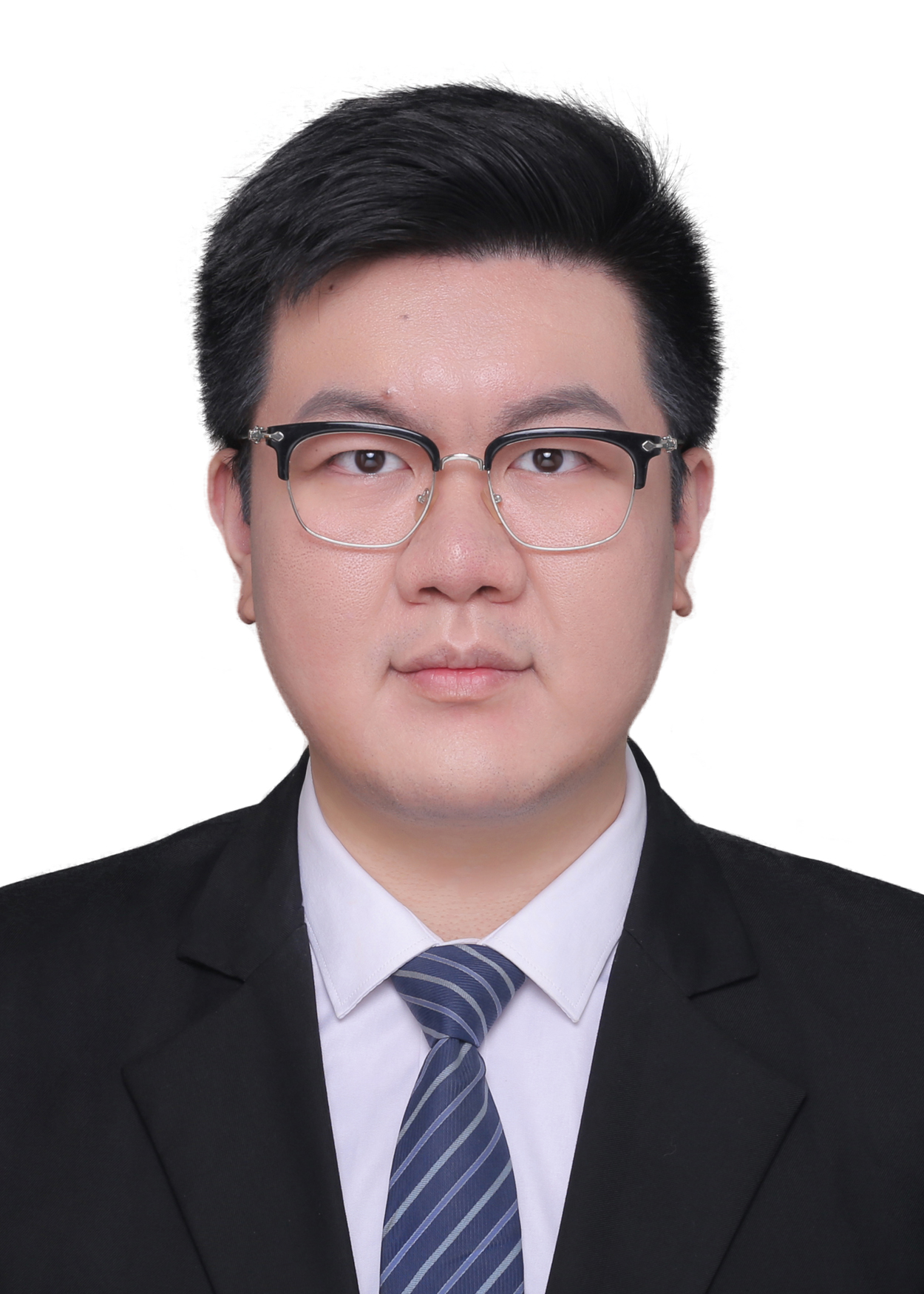}}]{Jincen Jiang}
is currently pursuing the Ph.D. degree with the National Centre for Computer Animation (NCCA), Bournemouth University, UK. He received the Master degree and the B.E. degree from the College of Information Engineering, Northwest A\&F University, China, in 2023 and 2020 respectively. His research interests include computer graphics, geometric modeling, and 3D deep learning.
\end{IEEEbiography}

\begin{IEEEbiography}[{\includegraphics[width=1in,height=1.25in,clip,keepaspectratio]{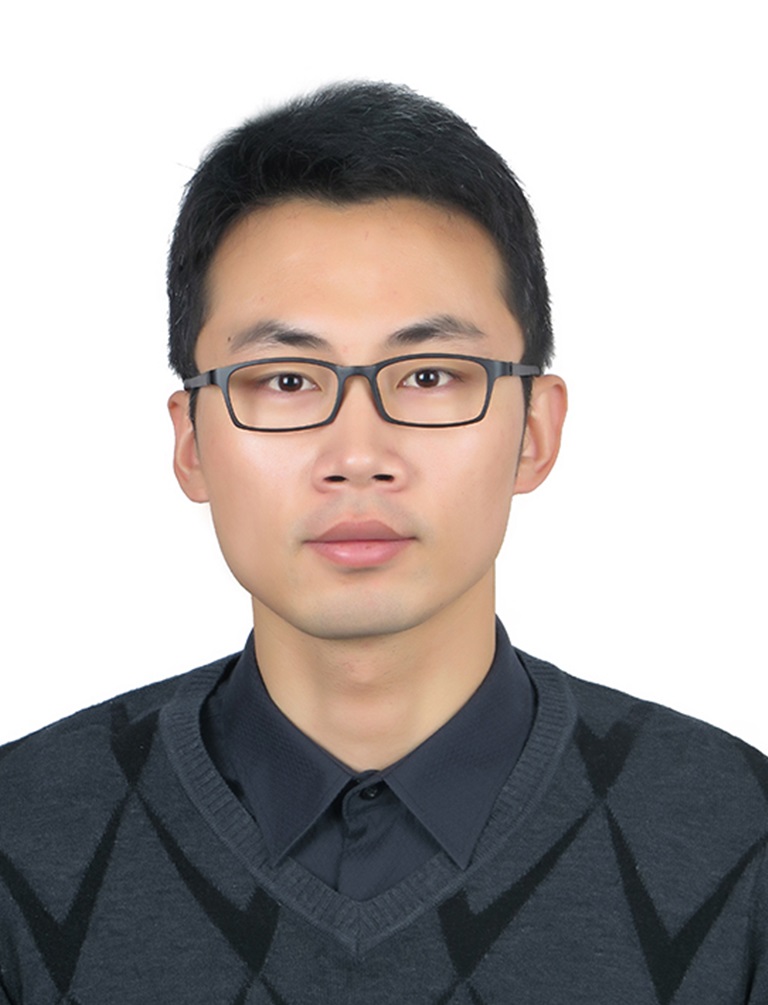}}]{Xuequan Lu}
(Senior Member, IEEE) is a Senior Lecturer at the Department of Computer Science and IT, La Trobe University, Australia. He spent more than two years as a Research Fellow in Singapore. Prior to that, he earned his PhD at Zhejiang University (China) in June 2016. His research interests mainly fall into the category of visual computing, for example, geometry modeling, processing and analysis, animation/simulation, 2D data processing and analysis. More information can be found at http://www.xuequanlu.com.
\end{IEEEbiography}

\begin{IEEEbiography}[{\includegraphics[width=1in,height=1.25in,clip,keepaspectratio]{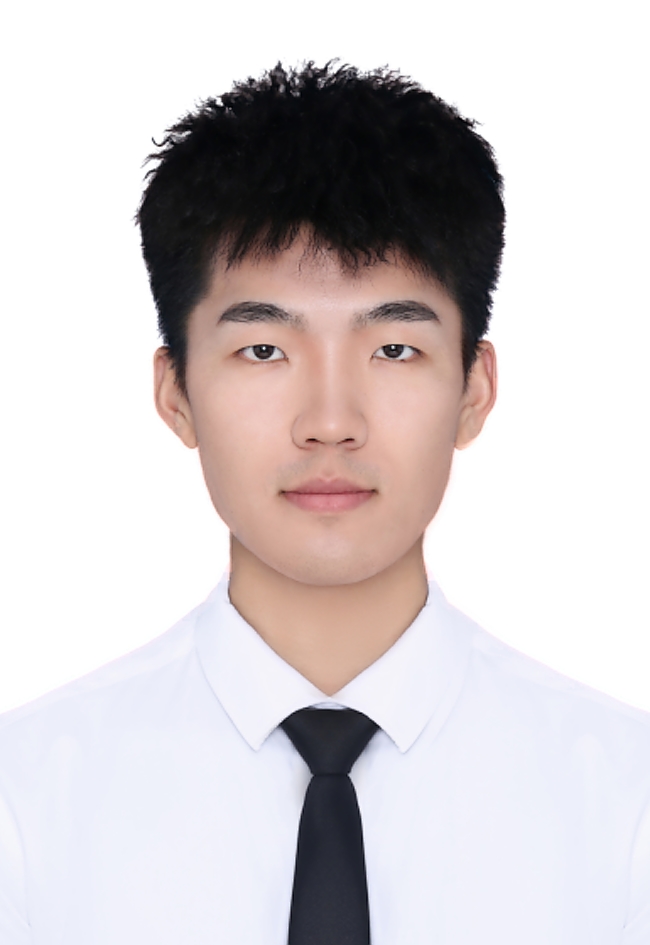}}]{Lizhi Zhao}
is currently working on the Ph.D. degree with the State Key Laboratory of Virtual Reality Technology and Systems, School of Computer Science and Engineering, Beihang University. He received his Master degree from the College of Information Engineering, Northwest A\&F University, China. His research interests mainly include computer graphics, virtual reality, and computer vision.
\end{IEEEbiography}

\begin{IEEEbiography}[{\includegraphics[width=1in,height=1.25in,clip,keepaspectratio]{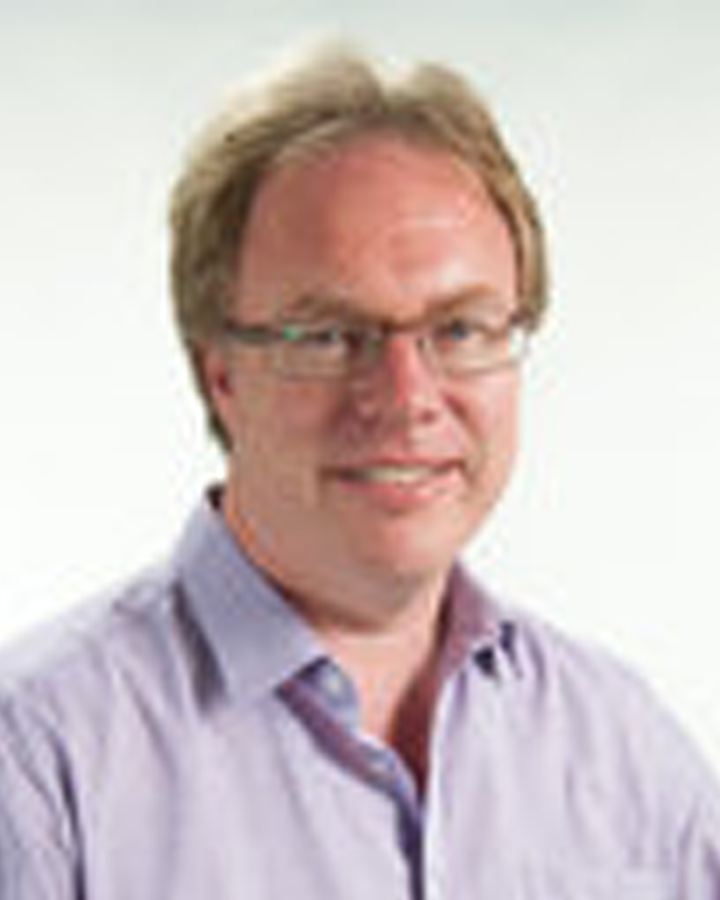}}]{Richard Dazaeley}
received his Ph.D. in Computer Science from the University of Tasmania in 2007. He is widely recognised for his pioneering work in multiobjective reinforcement learning publishing several highly cited papers that helped established the field. He has also organised the multiple workshops in the field. He is an associate professor of Computer Science at Deakin University (Geelong) where he is the Deputy Leader of the Machine Intelligence Lab. He is also interested in prudence analysis, natural language processing, data analytics and security. He has received multiple awards including Australian Educator of the Year in 2016 from the Australian Computer Society.
\end{IEEEbiography}

\begin{IEEEbiography}[{\includegraphics[width=1in,height=1.25in,clip,keepaspectratio]{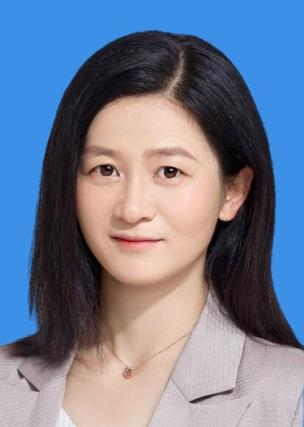}}]{Meili Wang}
(Member, IEEE) is a professor at College of Information Engineering, Northwest A\&F University. She received her Ph.D. degree in computer animation in 2011 at the National Centre for Computer Animation, Bournemouth University. Her research interests include computer graphics, geometric modeling, image processing, visualization and virtual reality.
\end{IEEEbiography}

\vfill

\end{document}